\documentclass[final,3p,times,authoryear]{elsarticle} 

\usepackage[dvipsnames,svgnames,x11names]{xcolor}

\usepackage[draft]{changes}
\usepackage[inline]{enumitem}
\definechangesauthor[color=blue]{PC}
\usepackage{lineno,hyperref}
\usepackage{amsmath,amssymb,amsfonts}
\usepackage{algorithmicx}
\usepackage{booktabs}
\newcommand{\concat}{\mathbin{\Vert}} 
\usepackage{booktabs}
\usepackage{tikz}
\usetikzlibrary{shapes.geometric, arrows.meta, positioning}
\usepackage{float}
\tikzstyle{block} = [rectangle, rounded corners, minimum width=3.5cm, minimum height=1.2cm, text centered, draw=black, fill=blue!10]
\tikzstyle{arrow} = [thick, ->, >=stealth]
\usepackage{anyfontsize}
\usepackage{adjustbox}
\usepackage{tablefootnote}
\usepackage{longtable}
\usepackage{tabularray}
\usepackage{caption}
\usepackage{subfigure}
\usepackage[font=small,labelfont=bf, figurename=Fig.]{caption} 

\usepackage{textcomp}
\usepackage{todonotes}
\usepackage{tabularx}
\usepackage{gensymb}
\usepackage[utf8]{inputenc}
\usepackage{mathtools, nccmath}
\usepackage{subcaption}
\usepackage{booktabs}
\usepackage{cleveref}
\usepackage{chemformula}
\usepackage[version=4]{mhchem}
\usepackage{booktabs, multirow} 
\usepackage{soul}
\usepackage{tablefootnote}
\modulolinenumbers[1]

\usepackage{algorithm}
\usepackage{algpseudocode}
\usepackage[authoryear]{natbib}
\definecolor{DarkGreen}{rgb}{0.2,0.5,0.2}
\hypersetup{
    colorlinks=true, 
    citecolor=DarkGreen,  
    linkcolor=black, 
    filecolor=black,
    urlcolor=blue    
}

\usepackage{setspace}

\makeatletter
\def\ps@pprintTitle{%
   \let\@oddhead\@empty
   \let\@evenhead\@empty
   \let\@oddfoot\@empty
   \let\@evenfoot\@empty}
\makeatother

\begin{document}
\begin{frontmatter}
\title{Surveillance Video-Based Traffic Accident Detection Using Transformer Architecture}

\author[mymainaddress]{Tanu Singh}

\author[mymainaddress]{Pranamesh Chakraborty
\corref{mycorrespondingauthor}}

\author[mymainaddress2]{Long T. Truong}

\cortext[mycorrespondingauthor]{Corresponding author\\
Pranamesh Chakraborty: pranames@iitk.ac.in,+91-512-259-2146}

\address[mymainaddress]{Department of Civil Engineering, Indian Institute of Technology Kanpur, Kanpur-208016, U.P., India}

\address[mymainaddress2]{School of Computing, Engineering and Mathematical Sciences, La Trobe University, Melbourne, Victoria-3086, Australia}

\begin{abstract}
Road traffic accidents represent a leading cause of mortality globally, with incidence rates rising due to increasing population, urbanization, and motorization. Rising accident rates raise concerns about traffic surveillance effectiveness. Traditional computer vision methods for accident detection struggle with limited spatiotemporal understanding and poor cross-domain generalization. Recent advances in transformer architectures excel at modeling global spatial–temporal dependencies and parallel computation. However, applying these models to automated traffic accident detection is limited by small, non-diverse datasets, hindering the development of robust, generalizable systems. To address this gap, we curated a comprehensive and balanced dataset that captures a wide spectrum of traffic environments, accident types, and contextual variations. Utilizing the curated dataset, we propose an accident detection model based on a transformer architecture using pre-extracted spatial video features. The architecture employs convolutional layers to extract local correlations across diverse patterns within a frame, while leveraging transformers to capture sequential-temporal dependencies among the retrieved features. Moreover, most existing studies neglect the integration of motion cues, which are essential for understanding dynamic scenes, especially during accidents. These approaches typically rely on static features or coarse temporal information. In this study, multiple methods for incorporating motion cues were evaluated to identify the most effective strategy. Among the tested input approaches, concatenating RGB features with optical flow achieved the highest accuracy at 88.3\%. The results were further compared with vision language models (VLM) such as GPT, Gemini, and Llava-Next-Video to assess the effectiveness of the proposed method. 
\end{abstract}

\end{frontmatter}
\section{Introduction}
Road traffic injuries rank among the leading causes of death worldwide, claiming approximately 1.19 million lives annually and causing 20 to 50 million non-fatal injuries \citep{WHO2023}. Although various efforts have been made to improve road monitoring—such as installing traffic surveillance at intersections—serious traffic violations remain common. These violations not only include over-speeding, but also reckless lane changing, running red lights, drunk driving, and distracted driving due to mobile phone use. It often leads to severe accidents, and unfortunately, many lives are lost due to delayed emergency medical response. One of the key reasons for this delay is the lack of immediate accident reporting to emergency services \citep{babu2022accident}. Early detection does more than just address the initial accident; it also helps prevent secondary crashes. When a problem on the road is identified quickly, traffic operators can respond right away by sending help, adjusting traffic controls, or warning drivers in real time. Therefore, early detection plays an important role in keeping the overall roadway environment safer.

Recent advancements in artificial intelligence (AI) have driven significant progress in computer vision. 
Advances in AI and increased computational power have enabled more accurate and efficient processing of visual data, leading to the development of intelligent video surveillance systems. These systems leverage advanced techniques, such as object detection, tracking, and event recognition, to support human operators in making timely, informed decisions. The continuous evolution of these technologies has led to their adoption in multiple surveillance applications, including border security, retail analytics, and emergency response.

Utilizing this technology in the field of traffic engineering to create an efficient traffic management system is crucial because transportation networks are constantly growing. The substantial research on traffic pattern analysis has been made possible by the widespread placement of cameras throughout urban areas and highways. Monitoring traffic dynamics, evaluating traffic behavior, tracking cars, and anomaly detection are important research topics \citep{hajri2022vision}.

Detecting anomalies on roads is inherently challenging due to various factors, including low-resolution images of distant vehicles, perspective-induced size variations, partial occlusions, and the presence of stationary cars in traffic jams. These challenges often result in incomplete or lost observations of target objects in video streams. So, we propose an approach to address these complexities effectively and accurately.

Although numerous approaches have been proposed and developed, traditional computer vision approaches for automated accident detection in surveillance videos suffer from limited spatiotemporal understanding and poor generalization across different domains. Recent advances in transformer architectures \citep{vaswani2017attention} have shown remarkable success in efficiently modeling global spatial-temporal dependencies and enabling parallel computation. These attributes have made them increasingly attractive for video understanding tasks. However, in traffic-surveillance applications, particularly those focused on accident detection and analysis, applying these models is still constrained by the limited availability of large-scale, diverse, and realistically representative datasets. Existing traffic accident video datset collections often lack sufficient variation in traffic conditions, camera-angle perspectives, accident types, and real-world complexity. Consequently, models trained on these datasets often exhibit limited robustness and generalization when applied in practical scenarios.

To address these limitations, we curated a dataset that has been developed to be both balanced and well-representative of real-world. This dataset includes a wide range of traffic scenes, vehicles, road infrastructures, and environmental conditions such as weather, lighting, and congestion levels. By incorporating a broad-range of accident and non-accident scenarios, the dataset more accurately represents the variability present in real-world traffic environments.

Building on this curated dataset, we propose a Transformer-based classification architecture for video-level classification using pre-extracted frame-level features. It considers both temporal and spatial features. Convolutional neural networks are used to extract the spatial characteristics, whereas  Transformers with positional embeddings and multi-head attention mechanisms are used to correlate the temporal features. Furthermore, most researchers overlook motion cues, even though they are essential for accurately capturing accidents, which unfold as continuous events. In our work, we explicitly incorporate these motion cues to enhance the understanding and interpretation of accident progression. We experimented with different approach to incorporate motion cues using optical flow to assess the model’s performance by integrating motion information with spatial visual information. 

Apart from the introduction, the structure of this paper is organised as follows. The Section~\ref{sec: liter} provides a comprehensive review of the existing literature. Section~\ref{sec: data} details the data collection process and the dataset used throughout the research. The detail of the proposed framework is described in Section ~\ref{sec: method}. Section~\ref{sec: results} consists of the results and analysis of the study. Finally, Section~\ref{sec: conclusion} summarises the findings of the paper and discusses potential directions for future research.

\section{Literature Review}
\label{sec: liter}
Over the past few years, deep learning has emerged as a foundational approach for traffic accident detection, with Convolutional Neural Networks (CNNs) playing a central role. The \cite{robles2021automatic} proposed frameworks that use CNNs to extract detailed spatial features and combine them with recurrent neural networks (RNNs), such as Long Short-Term Memory (LSTM) units, to model temporal dependencies and dynamic motion patterns across frames. This approach helps capture both appearance and temporal information from dashcam and surveillance videos, making real-time accident detection feasible.

Building on this approach, researchers have highlighted the importance of modeling object interactions within traffic scenes to detect and locate accidents. By combining deep CNNs for object feature extraction with spatiotemporal graph attention networks, models learn to understand how vehicles and other road users relate over time and space \citep{thakare2022object}. This highlights the advantage of graph-based temporal modeling in capturing complex patterns linked to accidents. In addition to spatial detection, real-time highway accident monitoring systems have been created that use YOLOv5 CNN-based object detectors combined with StrongSORT tracking. These methods use YOLOv5's effective single-stage detection to infer accidents by identifying unusual behaviors and anomalous vehicle stoppages through temporal analysis \citep{dharmadasa2023video}. However, the backbone of these approaches depends largely on the accuracy of object detection, which can be significantly affected by challenging lighting or noisy environments.

Parallel efforts have focused on leveraging high-performance CNN models like EfficientNet-B7 to make spatial features more robust in CCTV footage. By combining advanced pooling, normalization, and data augmentation, these models generalize better in surveillance settings, confirming that deep CNNs remain highly effective in detecting accidents in video \citep{singh2024efficientnet}. Further advancements include hybrid architectures that fuse CNN spatial feature extraction with bidirectional LSTMs for temporal modeling, as in modified Long-term Recurrent Convolutional Networks (LRCNs). These models classify various traffic incident types by effectively learning spatial-temporal patterns from video sequences, showcasing the continued effectiveness of CNN-RNN combinations for incident detection \citep{staby2023spatial}.

\cite{zhou2023appearance} proposed CNN architectures that address complex traffic scenarios with congested and occluded scenes by fusing appearance and motion information. Lightweight backbones like MobileNetV3 extract spatial features, while CNN-based optical flow learners capture motion cues. Temporal attention modules focus on critical frames, and auxiliary triplet loss improves feature robustness.

Despite these advances, CNN-RNN and graph-based models face inherent limitations in capturing long-range temporal dependencies and global contextual relationships due to the sequential nature of recurrent units and the local receptive fields of CNNs. They may also encounter scalability challenges and computational overhead as scene complexity increases. With their self-attention mechanisms, transformer-based architectures can overcome these issues by efficiently modeling global spatial-temporal dependencies and enabling parallel computation, thereby improving accident detection accuracy and robustness. Integrating transformers into these frameworks could address current shortcomings and enhance future traffic accident analysis systems.

The use of transformer architectures for video anomaly and accident detection has evolved significantly in recent years, demonstrating increasing effectiveness in capturing complex spatiotemporal dependencies and enabling real-time performance.

The fusion of convolutional neural networks with Vision Transformers (ViT) has further advanced accident detection from dashcam videos. By combining DenseNet for extracting local spatial features with ViT modules to capture temporal dynamics across frames, these models effectively bypass explicit modeling the spatiotemporal structure of video sequences via attention mechanisms. This hybrid architecture shows improved performance over traditional sequential models such as CNN-RNN and ConvLSTM \citep{hajri2022vision}.

Recent research \cite{yu2024fine} addresses the challenges of fine-grained accident detection in surveillance footage by developing datasets with diverse accident categories and lightweight models for edge device deployment. An enhanced detection framework based on You Only Look Once version 8 (YOLOv8) \citep{varghese2024yolov8} improves small object detection and integrates spatial attention modules. 

Lightweight autoencoder models that combine CNNs with transformer blocks have been introduced to denoise surveillance videos and enhance real-time anomaly detection. These systems encode complex spatiotemporal relationships using transformer architectures, reconstruct clean video frames, and identify deviations that indicate anomalies. This approach demonstrates robustness in scenarios with significant video noise \citep{roka2023d}. Recent advancements include unsupervised frameworks that use vision transformers with convolutional spatiotemporal attention blocks, effectively capturing both local and global dependencies in surveillance videos. The framework enhances robustness to diverse real-world scenarios while remaining scalable for modern large-scale surveillance applications \citep{habeb2024enhancing}. However, while the model generates frame-level anomaly scores, it does not directly produce video-level anomaly predictions.

Most of the existing methods generate frame-level anomaly scores, making them effective for detecting anomalies at the frame level. However, they cannot directly produce video-level anomaly predictions, relying instead on post-hoc aggregation of frame-level scores, which may be suboptimal for accident detection that requires holistic video-level decisions instead of frame-level.

Motion cues are essential for understanding dynamic scenes, especially in the case of an accident. Motion cues capture pixel-level changes and object movements that static features or general temporal information cannot represent. Despite their importance, the majority of existing studies have largely overlooked the use of motion cues in the accident detection framework, often relying solely on static features or coarse temporal information, leaving a critical gap in accurately capturing dynamic scene changes and object interactions. In this study, we leverage optical flow to explicitly encode motion between consecutive frames and incorporate it through multiple strategies, including fusion with RGB frames and video-level integration to emphasize regions with significant movement. This allows the model to learn rich spatio-temporal representations, capturing nuanced dynamics and complex interdependencies.

Inspired by the architectural fusion of convolutional networks and transformer models, we propose a framework to extract spatial features using CNNs and transformer-based architecture to model the temporal dynamics of the traffic scene to detect an accident in a surveillance video. Moreover, since the existing dataset lacks sufficient diversity and real-world complexity, we curated a dataset that addresses the limitations present in existing datasets. Additionally, we also investigated different ways to integrate motion dynamics cues along with spatial information to enhance the understanding and interpretation of accident progression by the model.

\section{Dataset}
\label{sec: data}

The development of a transformer-based accident detection model relies heavily on the availability of a large, diverse, and high-quality dataset. To support robust model training and evaluation, this study utilised traffic accident and non-accident video data directly obtained from CCTV surveillance cameras. The use of CCTV footage provides a significant advantage, as it naturally captures a wide variety of real-world traffic scenarios.. These include different traffic densities (from free-flowing conditions to heavy congestion), roadway types such as highways, urban roads, and signalized or unsignalized intersections, multiple camera angles, various times of day (from early morning to late night), diverse lighting conditions, and multiple weather patterns including clear skies, rainfall, and snowfall. Together, these factors contribute to a rich and heterogeneous dataset.

The accident-related CCTV videos were curated from multiple sources, including YouTube and publicly available benchmark datasets such as IEEE DataPort (Adewopo et al., 2023) and the AICity Challenge dataset (Naphade et al., 2021). Relevant clips were identified using keywords like “traffic accident” and “road collision,” specifically focusing on surveillance-style viewpoints. In contrast, the non-accident CCTV clips were collected from live traffic surveillance feeds in Kanpur city (India) and Florida (USA). These non-accident clips capture regular traffic flow at different times of the day morning, afternoon, and evening thereby offering natural variability in lighting, density, and vehicle interactions. Before finalising the dataset, we carefully preprocessed all videos for consistency with a pre-existing dataset. Each clip was trimmed to a similar length to match the duration of the typical accident videos, making sure both accident and normal clips were comparable, concise segments. Whenever possible, we selected accident clips that showed a few seconds before the crash, the moment of impact, and the immediate aftermath. This gives the model important context about how accidents unfold.

The final dataset consists of 1000 clips, with accident and non-accident classes represented equally. The average duration of videos is 7.5 seconds, which is long enough to show a developing event but short enough for efficient processing. The balanced composition of the dataset helps prevent class bias during model training. Figure 1 provides sample images from this diverse collection, illustrating the variety of scenes and conditions included.

\begin{figure}[h]
    \centering
    \includegraphics[width=1\linewidth]{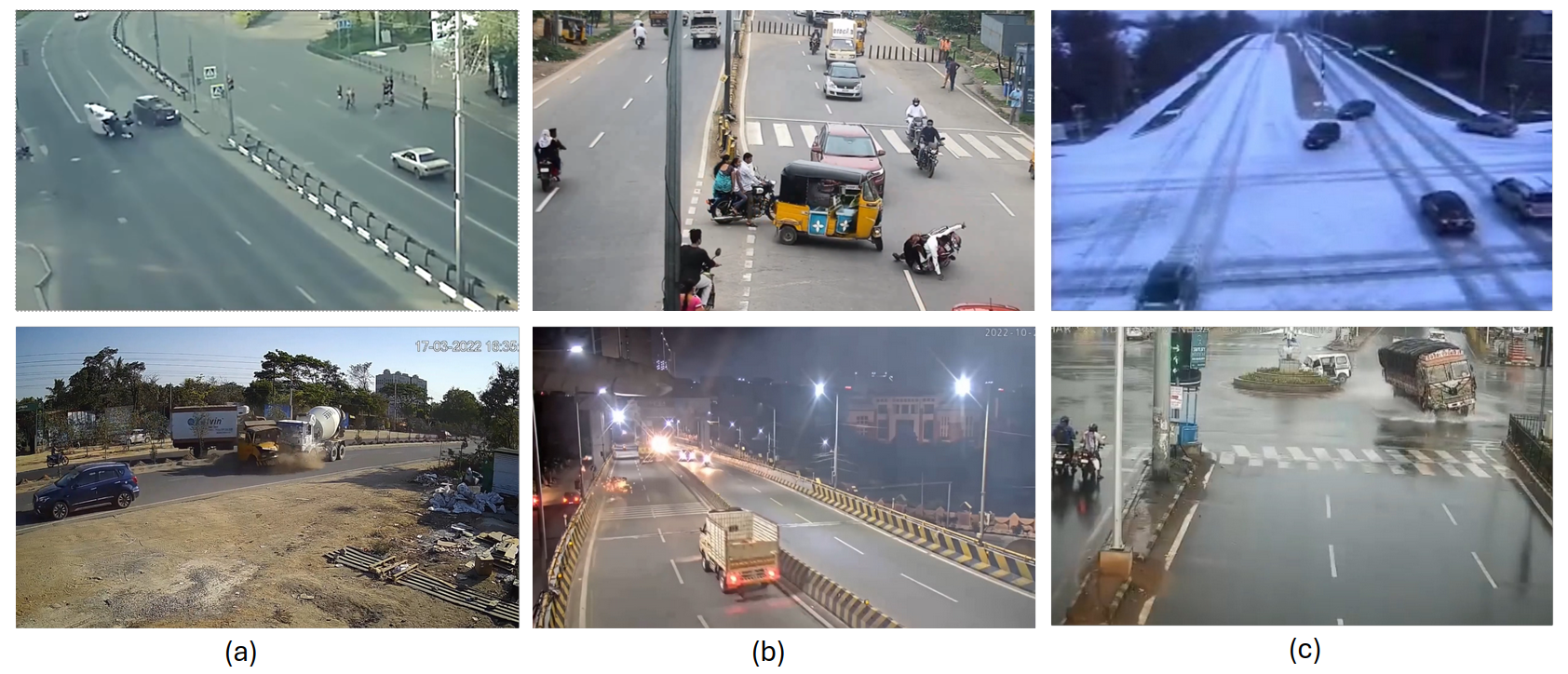}
    \caption{Sample frames from the accident video dataset, illustrating diversity in (a) camera angle, (b) time of day, and (c) weather conditions.}
    \label{dataset}
    \end{figure}

\section{Methodology}
\label{sec: method}
This section presents the framework proposed to develop an accident detection model based on a transformer architecture using pre-extracted spatio-temporal video features. The model architecture combines sequence modeling capabilities of transformers with efficient feature extraction, enabling accurate classification of accident and non-accident scenarios. Therefore, the proposed framework can be divided into two phases: Feature extraction and Transformer-based detection Model.
\subsection{Feature extraction} 
\label{subsec: feature}

This phase of the proposed pipeline processes videos by sampling frames, extracting features using a pretrained convolutional neural network (CNNs).

To ensure computational efficiency while retaining representative visual content, we uniformly sample every fifth frame from each video. 
We extract features for each sampled frame using pre-trained convolutional models. The final classification layer is removed, resulting in a convolutional feature extractor that outputs a feature vector for each frame. The input frames are resized to 224×224 pixels and normalized using the standard ImageNet mean and standard deviation. This process yields robust, high-level semantic representations of the visual content. 

Several CNN architectures, including ResNet \citep{resnet}, DenseNet \citep {densenet}, and EfficientNet \citep{efficientnet}, were evaluated to compare their effectiveness in feature extraction. Among all the models, ResNet consistently outperformed the others, producing more discriminative and robust feature representations. Therefore, ResNet was selected for feature extraction in this study.

\subsection{Detection Model Architecture}

We propose a Transformer-based sequence classification architecture for video-level classification using pre-extracted frame-level features. The model is designed to process variable-length sequences and predict a video-level class label. It follows the steps illustrated in Figure \ref{flowchart}.

Each input video is represented as a sequence of frame-level feature vectors extracted using a pre-trained convolutional neural network. Each frame is represented as a vector \( \mathbf{x}_t \in \mathbb{R}^{2048} \), and a video of length \( T \) is represented as a sequence \( \{\mathbf{x}_1, \mathbf{x}_2, \ldots, \mathbf{x}_T\} \in \mathbb{R}^{T \times 2048} \). 
To map the input features to a suitable embedding space for the transformer, we use a linear projection:

\[
\mathbf{z}_t = \mathbf{W}_p \mathbf{x}_t + \mathbf{b}_p, \quad \mathbf{z}_t \in \mathbb{R}^{d_{\text{model}}}
\]

Transformers have no inherent sense of order, since they process input as a set of vectors rather than a sequence. To provide information about the position of each frame within a video, a positional encoding is added to the input embeddings. This positional encoding uses sine and cosine functions of varying frequencies to create a unique positional vector for each time step. These encodings are added to the input features so that the model can learn patterns over time (for example, motion or change in scene). 
\[
\mathbf{z}'_t = \mathbf{z}_t + \text{PE}(t)
\]

Next, the encoded sequence \( \mathbf{Z}' = \{\mathbf{z}'_1, \ldots, \mathbf{z}'_T\} \) is passed through a stack of Transformer encoder layers. Each layer consists of multi-head self-attention, followed by position-wise feedforward layers, layer normalization, and dropout. The Transformer output is:

\[
\mathbf{H} = \text{TransformerEncoder}(\mathbf{Z}', \text{mask})
\]

where the mask prevents attention to the padding positions.

\tikzstyle{block} = [
    rectangle,
    rounded corners,
    draw=black,
    thick,
    fill=gray!10,
    text centered,
    text width=6cm,
    minimum height=1cm
]
\tikzstyle{arrow} = [thick, -{Latex[length=3mm, width=2mm]}]

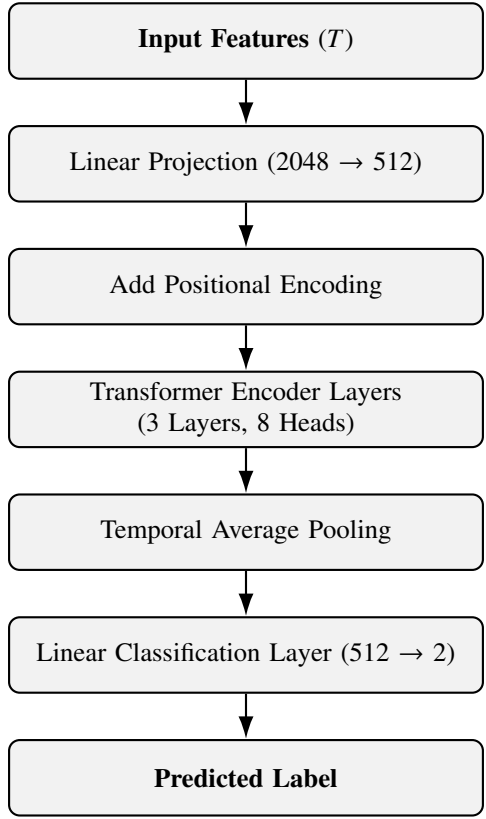
\begin{figure}[h]
\centering
\begin{tikzpicture}[node distance=0.9cm] 

\node (input) [block] {\textbf{Input Features} ($T$)};
\node (proj) [block, below=6mm of input] {Linear Projection (2048 → 512)};
\node (posenc) [block, below=6mm of proj] {Add Positional Encoding};
\node (transformer) [block, below=6mm of posenc] {Transformer Encoder Layers \\ (3 Layers, 8 Heads)};
\node (pool) [block, below=6mm of transformer] {Temporal Average Pooling};
\node (classifier) [block, below=6mm of pool] {Linear Classification Layer (512 → 2)};
\node (output) [block, below=6mm of classifier] {\textbf{Predicted Label}};

\draw [arrow] (input) -- (proj);
\draw [arrow] (proj) -- (posenc);
\draw [arrow] (posenc) -- (transformer);
\draw [arrow] (transformer) -- (pool);
\draw [arrow] (pool) -- (classifier);
\draw [arrow] (classifier) -- (output);

\end{tikzpicture}

\caption{Overview of the proposed Transformer-based detection model. Input features are linearly projected, positionally encoded, and processed through Transformer encoders. The temporal representations are averaged and classified to generate the final prediction.}
\label{flowchart}
\end{figure}

Due to variability in video lengths, feature sequences are padded to ensure uniform sequence lengths within each batch. A masking mechanism is introduced to differentiate between actual and padded frames. This mask is used within the transformer model to prevent attention operations on padded positions, thereby preserving the integrity of temporal modeling.

Next, we applied average pooling over the temporal dimension to obtain a fixed-size vector representation of the video:

\[
\mathbf{h}_{\text{video}} = \frac{1}{T} \sum_{t=1}^{T} \mathbf{H}_t
\]

The pooled representation is passed through a linear classifier to obtain the logits for each class:

\[
\mathbf{y} = \mathbf{W}_c \mathbf{h}_{\text{video}} + \mathbf{b}_c
\]
where \( \mathbf{y} \in \mathbb{R}^{C} \) and \( C = 2 \) for binary classification.


The model is trained using cross-entropy loss:

\[
\mathcal{L} = -\sum_{i=1}^{C} y_i \log(\hat{y}_i)
\]

where \( y_i \) is the true label and \( \hat{y}_i \) is the predicted probability after softmax.

The hyperparameters are mentioned in Table \ref{tab:HYPERPARAMETER} with their corresponding values.
\begin{table}[h]
\centering
\caption{Hyperparameters used in transformer-based model for accident detection}
\begin{tabular}{ll}
\hline
\textbf{Component}          & \textbf{Value}            \\
\hline
Input Feature Dim           & 2048                      \\
Embedding Dim               & 512                       \\
Transformer Layers          & 3                         \\
Attention Heads             & 8                         \\
Dropout Rate                & 0.1                       \\
Sequence Pooling            & Average pooling           \\
Classification              & Linear (512 $\rightarrow$ 2) \\
Loss Function               & Cross-Entropy             \\
\hline
\end{tabular}

\label{tab:HYPERPARAMETER}
\end{table}

\subsection{Different input modalities}

\begin{figure}[h]
    \centering
    \includegraphics[width=1\linewidth]{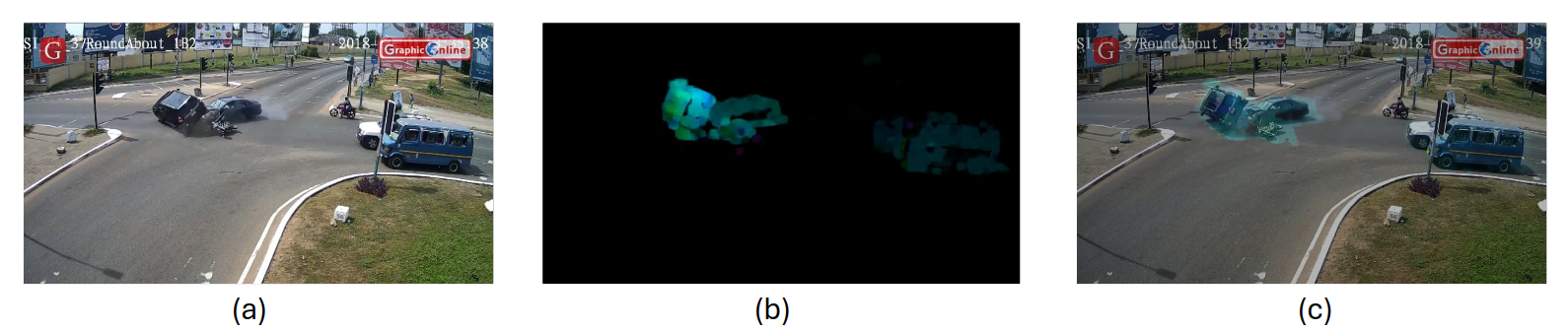}
    \caption{Illustration of different input modalities (a) Original RGB Frame (b) Optical flow frame (c) optical flow overlaid frame}
    \label{3exp}
    \end{figure}

To thoroughly evaluate the model and identify which feature best captures the abrupt change in vehicle motion during the accident, we conducted the study using four different input modalities. Each follows the same processing pipeline consisting of feature extraction followed by transformer model-based detection. These differ only in the type of input provided to the pipeline:

\begin{itemize}
    \item \textbf{Input Methods 1: Original RGB Video Input}-
    In the first method, the raw RGB video frames were used as input. Each video sequence was processed frame-by-frame to extract spatial features that capture color, texture, and visual appearance. These features were then passed to the detection model for training and inference. This setup mainly focuses on appearance-based information.

    \item \textbf{Input Methods 2: Optical Flow Video Input}-
    In the second method, we used the optical flow representation of the same videos as input. Optical flow encodes the motion between consecutive frames, thus emphasizing the temporal dynamics and motion patterns of objects. This representation allows the model to focus on motion cues rather than static visual information. The optical flow fields were first computed as shown in Figure \ref{3exp} (b) and then used for feature extraction before being fed into the detection model.

    \item \textbf{Input Methods 3: Optical Flow Overlaid Video Input}-
    In the third configuration, the optical flow overlaid video was used as input. The optical flow was blended with the original RGB frames to generate an optical flow overlaid video as shown in Figure \ref{3exp}(c). The resulting input emphasizes motion patterns while retaining appearance information from the original video. Then, further, the same pipeline was used- feature extraction and detection model.

    \item \textbf{Input Methods 4: Features: RGB concatenated Optical flow}-
    In the fourth method, the features extracted from RGB frames and Optical flow frames were processed independently to capture distinct aspects of the input data. After extraction, the two feature sets were concatenated along the feature dimension to form a comprehensive representation that integrates information from both. This fused feature vector was fed into the detection model, allowing the model to utilize the enriched, multi-source feature space for improved detection performance.

\end{itemize}

All methods employ identical feature extraction architectures and detection model parameters to ensure a fair comparison of different experiments. The results of these experiments provide insights into how spatial (RGB) and temporal (optical flow) information contribute to overall detection performance.

\section{Results and analysis}
\label{sec: results}
In this section, we evaluate our model on the proposed dataset with different input approach. Accident detection systems demand highly robust models due to the unpredictable nature of accidents. To ensure the model can handle real-world scenarios, we need to evaluate it using a sufficiently large and varied dataset. Therefore, we split the dataset into a 70:30 ratio for the training and testing sets, respectively.

To assess the performance of the proposed model, several standard classification metrics were employed, namely \textit{accuracy}, \textit{precision}, \textit{recall}, and the \textit{F1-score}. These metrics provide a comprehensive understanding of the model's ability to classify samples and handle class imbalances.

The following metrics were computed :

\begin{itemize}
    \item {Accuracy:} measures the overall proportion of correctly classified samples.
    \[
    \text{Accuracy} = \frac{TP + TN}{TP + TN + FP + FN}
    \]
    
    \item {Precision:} quantifies the proportion of positive predictions that are actually correct.
    \[
    \text{Precision} = \frac{TP}{TP + FP}
    \]
    
    \item {Recall:} also known as sensitivity, measures the proportion of true positives correctly identified by the model.
    \[
    \text{Recall} = \frac{TP}{TP + FN}
    \]
    
    \item {F1-score:} represents the harmonic mean of precision and recall, providing a balanced measure between the two.
    \[
    \text{F1} = 2 \times \frac{\text{Precision} \times \text{Recall}}{\text{Precision} + \text{Recall}}
    \]
\end{itemize}

These metrics collectively offer a reliable evaluation of the model’s classification performance. The positive class was considered as label ``1'' throughout the analysis.

\begin{table}[h]
\centering
 \caption{Performance metrics for different experiments on the proposed dataset}
\begin{tabular}{@{}llcccccc@{}}
\toprule
\textbf{Features extracted} & \textbf{Accuracy} & \textbf{Precision} & \textbf{Recall} & \textbf{F1 Score}  \\ \midrule

1. RGB & 0.820 & 0.875 & 0.747 & 0.806 \\

2. Optical flow  & 0.873 & 0.918 & 0.820 & 0.866  \\

3. Optical flow overlaid video & 0.863 & 0.876 & 0.847 & 0.861 \\

4. RGB $ \concat $ Optical flow  & 0.883 & 0.881 & 0.887 & 0.884  \\
\bottomrule
\end{tabular}
\label{tab:perf}
\end{table}

Table \ref{tab:perf} summarizes the performance of different feature extraction approaches evaluated using accuracy, precision, recall, and F1 score. The feature extraction using original videos yielded the lowest performance, with an accuracy of 0.820 and an F1 score of 0.806, indicating that the frame features alone were insufficient for capturing the key patterns required for accurate classification. When additional motion information was incorporated into the feature set, there was a consistent improvement across all performance metrics. Method 2 achieved an accuracy of 0.873 and an F1 score of 0.866, showing a substantial improvement over the baseline. Method 3 maintained balanced performance (accuracy 0.863, F1 0.861) with slightly better recall, suggesting enhanced sensitivity to relevant instances. The best results were obtained using Method 4, which achieved an accuracy of 0.883, precision of 0.881, recall of 0.887, and F1 score of 0.884. This demonstrates that incorporating richer or more informative features significantly improves the model’s ability to generalize and accurately identify relevant patterns.

\begin{figure}[h]
    \centering
    \includegraphics[width=1\linewidth]{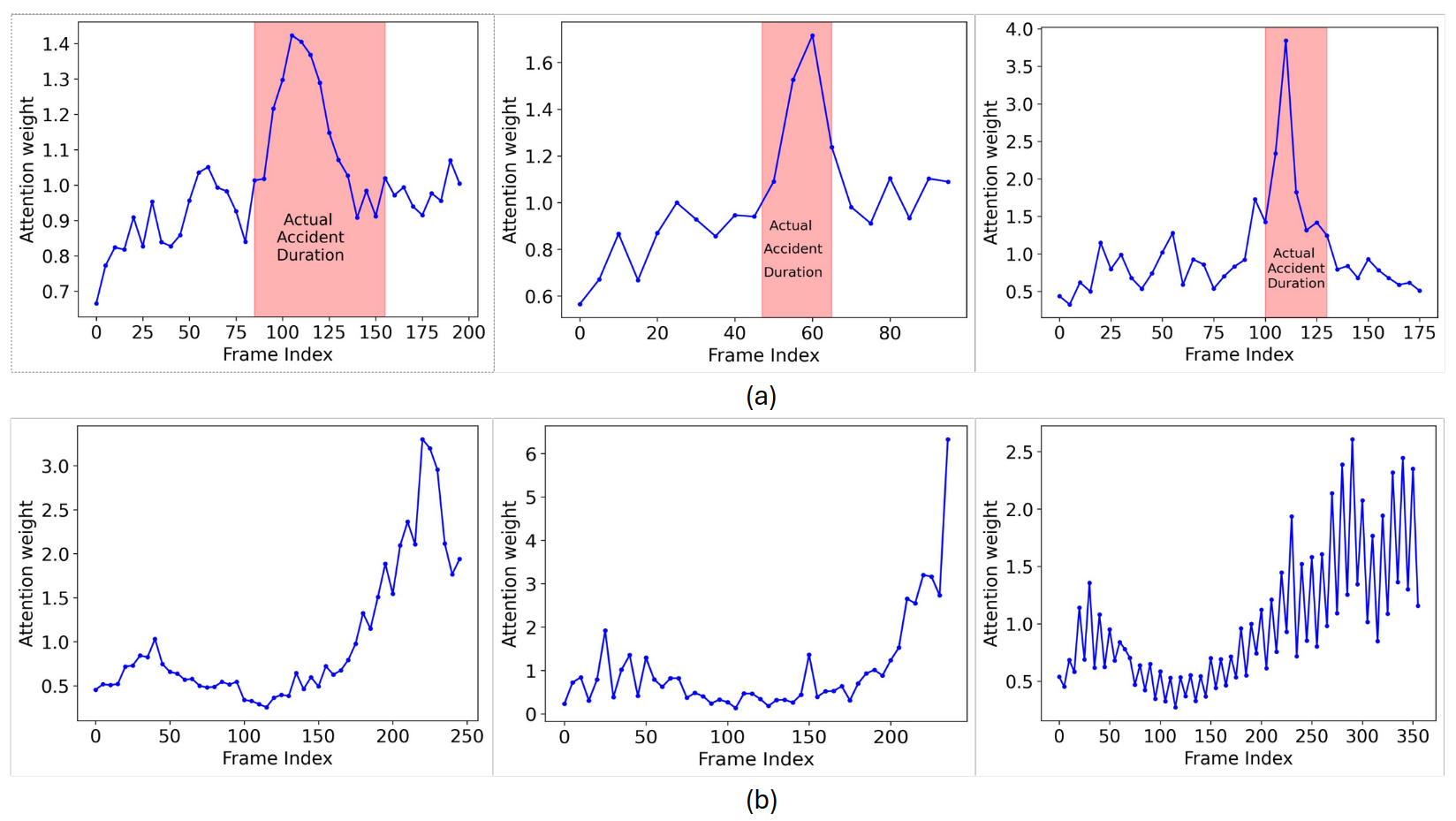}
    \caption{Frame-wise attention distributions for (a) true-positive cases and (b) true-negative cases. }
    \label{attention map}
    \end{figure}

Figure \ref{attention map} shows the attention weights used by the model in detecting an accident. Figure \ref{attention map} (a) shows the samples of True positives, whereas (b) shows weights for true negatives. A distinct pattern was observed: accident videos tend to generate concave graph trends, whereas non-accident videos typically show convex trends. For accident videos, the peaks are for the duration reflecting the time span of the accident in a video, represented as a red shaded region in Figure \ref{attention map} (a).

\begin{figure}[h]
    \centering
    \includegraphics[width=1\linewidth]{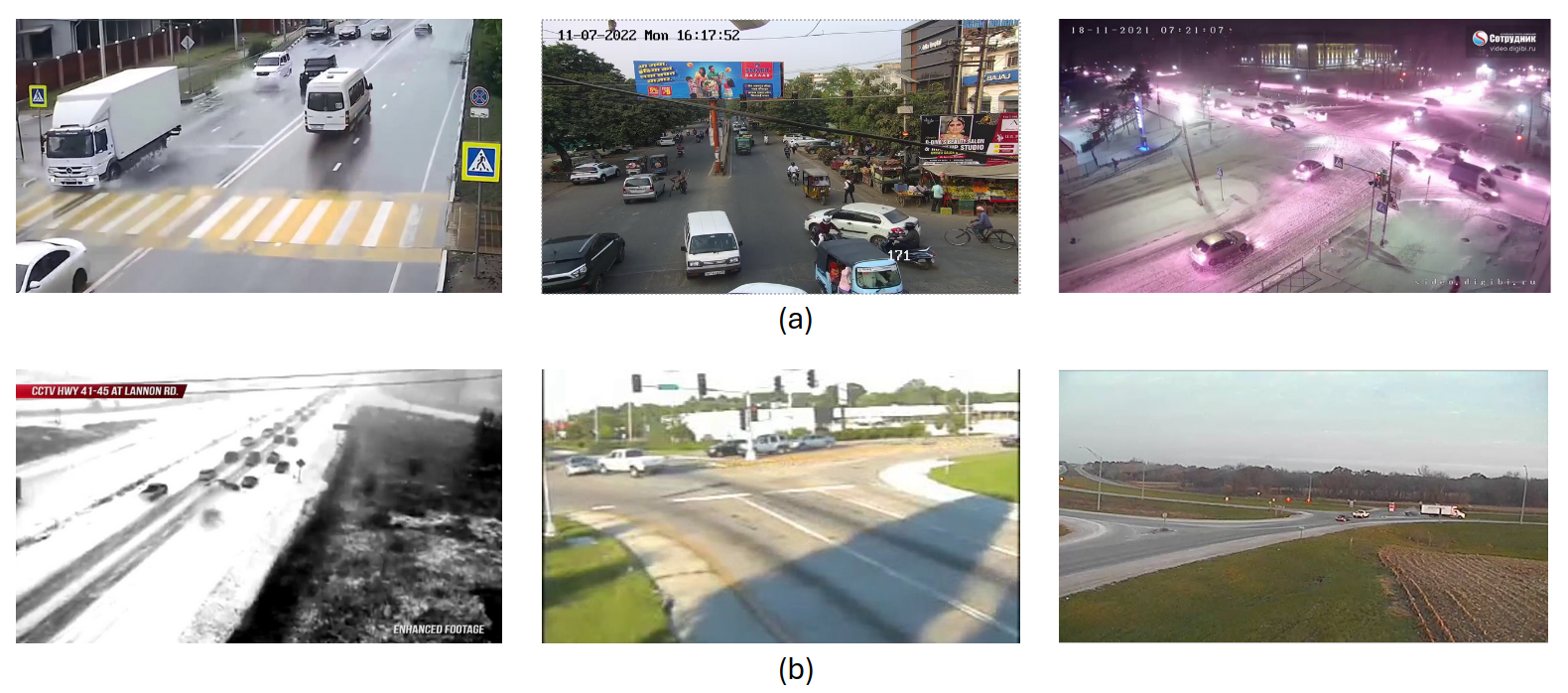}
    \caption{ Misclassified samples by the model: (a) False positive samples (b) False negative samples }
    \label{misclassification}
    \end{figure}

Figure \ref{misclassification} shows some examples of misclassifications, where (a) shows an example of frames when the model has misclassified the videos as accidents due to different reasons, such as the close proximity of vehicles and headlight glare. Sometimes, the camera’s visibility is hindered by weather conditions, leading to incorrect classification. Whereas, Figure \ref{misclassification} (b) represents some of the sample frames where the model has misclassified videos as non-accident.

To compare the performance of different methods, we conducted comparison experiments on our test dataset. Table \ref{tab:performance} represents the comparison of the result obtained from different state-of-the-art Vision Language Models (VLMs) such as GPT-5 \citep{gpt5.1}, Gemini \citep{google_gemini}, and Llava-Next-Video \citep{llavanextvideo}. VLMs were queried using the prompt: \textit{"Is there any traffic accident/crash in the video. Write Yes or No"} to generate the detection results for each video. From Table \ref{tab:performance}, it can be inferred that Gemini achieves higher performance across the evaluated methods, while GPT demonstrates a comparable result to our method. On the other hand, LLaVA-Next-Video, which is an open-source VLM model, underperformed relative to our model. Many of the better-performing VLMs are closed-source and require substantial computational resources, whereas our method is more accessible and cost-effective. These results indicate that our method achieves a favorable balance between competitive performance and practical usability, highlighting its potential as a viable option. We have also reproduced the method proposed by \cite{wsal} on our test data to compare the results. The results clearly indicate that our proposed approach significantly outperforms the existing method by a substantial margin. Overall, our proposed accident detection framework, which integrates motion cues, achieved promising results and demonstrates strong potential for real-world traffic monitoring applications.

\begin{table}[H]
\centering
 \caption{Comparision of Performance metrics on test data using different methods}
\begin{tabular}{@{}llcccccc@{}}
\toprule
\textbf{Method} & \textbf{Accuracy} & \textbf{Precision} & \textbf{Recall} & \textbf{F1 Score}  \\ \midrule

Gpt-5  & 0.890 & 0.976 & 0.800 & 0.879 \\

Gemini  & 0.903 & 0.929 & 0.873 & 0.900 \\

Llava-Next-Video & 0.660 & 0.632 & 0.767 & 0.693 \\

\cite{wsal}    & 0.647 & 0.615 & 0.787 & 0.690 \\
Proposed model  & 0.883 & 0.881 & 0.887 & 0.884 \\
\bottomrule
\end{tabular}
\label{tab:performance}
\end{table}

\section{Conclusion}
\label{sec: conclusion}
Road accident detection through traffic surveillance systems is essential for enhancing traffic monitoring, decreasing emergency response time, and improving road safety. In this paper, we present a hybrid method for accident detection that exploits both spatial and temporal features to capture the intricate dynamics of traffic incidents. Convolutional neural networks extract spatial attributes, while Transformers with positional embeddings and multi-head attention correlate temporal features. 
For this study, we curated a surveillance-based accident dataset representing diverse traffic scenarios, lighting conditions, and accident types. The dataset was meticulously constructed to be balanced and reflective of real-world conditions, then divided into training and testing sets to ensure rigorous model evaluation. Additionally, we explicitly integrate motion cues to advance the assessment and interpretation of accident progression. We explored multiple techniques for incorporating motion cues with optical flow to evaluate the model’s performance in fusing motion with spatial information. Among the tested input strategies, concatenating RGB features with optical flow yielded the highest accuracy at 88.3\%. Results were also benchmarked against VLM, including GPT, Gemini, and Llava-Next-Video, to evaluate the proposed method’s effectiveness. Overall, results are promising, and the framework demonstrates strong potential. We also recognize limitations in accurately detecting entities involved in accidents. Addressing these issues, our future work will focus on generating comprehensive accident descriptions alongside detection for more reliable outcomes.

\bibliography{bibliography}
\end{document}